\pdfoutput=1
\documentclass[11pt]{article}

\usepackage{ACL2023}

\usepackage{times}
\usepackage{latexsym}
\usepackage{caption}

\usepackage[T1]{fontenc}
\usepackage{booktabs,tabularx}

\usepackage[utf8]{inputenc}

\usepackage{microtype}

\usepackage{inconsolata}
\usepackage{array,multirow}

\usepackage{booktabs}
\usepackage{tabularx}
\usepackage{xcolor}
\usepackage{soul}

\newcommand{\hlc}[2][yellow]{{%
    \colorlet{foo}{#1}%
    \sethlcolor{foo}\hl{#2}}%
}

\usepackage[inline]{enumitem}
\usepackage[utf8]{inputenc} % allow utf-8 input
\usepackage[T1]{fontenc}    % use 8-bit T1 fonts
\usepackage[english]{babel}
\usepackage{hyperref}       % hyperlinks
\usepackage{url}            % simple URL typesetting
\usepackage{amsfonts}       % blackboard math symbols
\usepackage{nicefrac}       % compact symbols for 1/2, etc.
\usepackage{microtype}      % microtypography
\usepackage{xcolor}         % colors
\usepackage{booktabs}
\usepackage{amsmath}
\usepackage{multirow}
\usepackage{graphicx}
\usepackage{enumitem}
\usepackage{subcaption}
\usepackage{caption}
\usepackage{rotating}
\usepackage[normalem]{ulem}
\usepackage{amssymb}
\usepackage{colortbl}
\usepackage{linguex}

\usepackage{siunitx} 
\usepackage{multirow, makecell}
\usepackage{pifont}% http://ctan.org/pkg/pifont
\usepackage{siunitx} 
\usepackage{multirow, makecell}
\usepackage{tabularx}% added for table design
\usepackage[all]{nowidow}
\usepackage{appendix}
\usepackage{tikz}
\usetikzlibrary{decorations.pathreplacing,positioning,calc}
\definecolor{darkgreen}{rgb}{0.0, 0.42, 0.24}
\definecolor{green}{RGB}{112, 173,71}
\definecolor{blue}{RGB}{68, 114,196}
\definecolor{orange}{RGB}{237, 125,49}
\definecolor{red}{RGB}{202, 54,49}
\definecolor{yellow}{RGB}{222,194, 142}

\newcommand{\sbert}{\textsc{SBERT}\xspace}

\preto\tabular{\setcounter{magicrownumbers}{0}}
\newcounter{magicrownumbers}

\usepackage{enumitem}
\setlist{nolistsep,leftmargin=*}
\title{Keep It Private: Unsupervised Privatization of Online Text}

\author{Calvin Bao,  Marine Carpuat \\
\texttt{\{csbao, marine\}@umd.edu} \\
  University of Maryland, College Park \\
}

\begin{document}
\maketitle

% \nolinenumbers

\begin{abstract}

% \mc{Authorship obfuscation techniques hold the promise of helping people protect their privacy in online communication by automatically rewriting text to hide the identity of the original author. However, obfuscation has only been studied in narrow settings in the NLP literature and has primarily been addressed with character level edit operations that lead to unnatural outputs. In this work, we introduce an automatic text privatization model that fine-tunes a large language model via reinforcement learning to produce rewrites that balance soundness, sense and privacy. We evaluate it extensively on a large-scale test set of English Reddit posts by 68k authors, and study how the text length and adversarial authorship detection strategy impact its performance.}

Authorship obfuscation techniques hold the promise of helping people protect their privacy in online communications by automatically rewriting text to hide the identity of the original author. However, obfuscation has been evaluated in narrow settings in the NLP literature and has primarily been addressed with superficial edit operations that can lead to unnatural outputs. In this work, we introduce an automatic text privatization framework that fine-tunes a large language model via reinforcement learning to produce rewrites that balance soundness, sense, and privacy. We evaluate it extensively on a large-scale test set of English Reddit posts by 68k authors composed of short-medium length texts. We study how the performance changes among evaluative conditions including authorial profile length and authorship detection strategy. Our method maintains high text quality according to both automated metrics and human evaluation, and successfully evades several automated authorship attacks.

 % We show that our method jointly optimizes for those notions in the resulting texts and highlight weaknesses of obfuscation models in the evaluative conditions.  \mc{replace last sentence with something more concrete} 

%These techniques have usually been optimized at the character- and word- level, with the sole goal of fooling a set of authorship classifiers.
% Model submissions to the PAN obfuscation series vary from strictly rule-based stylometric approaches to autoencoder and transformer-based, while some work propose using different tasks (NMT or paraphrasing)

%We additionally find that performance in verification setting depends on length, wher
% efficient, distance-based rewards using an unsupervised reinforcement learning framework proposed by \cite{laban-etal-2021-keep}. We design rewards to jointly optimize for the obfuscation task in terms of sense and privacy. Chiefly, we leverage Learnt Universal Authorship Representations (LUAR) \cite{rivera-soto-etal-2021-learning} to enable privacy, and SBERT representations \cite{reimers2019sentencebert} to enable adequacy. Finally, we evaluate our obfuscation model in a way that considers privacy in both the verification and attribution settings on Reddit comments and Fanfiction data from the PAN 2020 Verification shared task \footnote{https://pan.webis.de/clef20/pan20-web/author-identification.html}.

% Nice to have, but if no time, remove
% Moreover, we introduce an evaluative task to obfuscation with the idea of ``de-obfuscation'' to assess the ease of reverse engineering obfuscated text.

\end{abstract}

\section{Introduction}

Maintaining privacy is crucial to allow everyone's participation in online communities. This is a key motivation for platforms such as Reddit that let users contribute pseudonymously. While anonymity is sometimes viewed as a cause of abuse, there is also clear evidence that de-anonymizing online comunication can harm ``queer people, sex workers, activists, researchers, journalists, and persons holding combinations of these identities'' \cite{afsaneh-2021-why}.
However simply using a pseudonym rather than one's legal name does not guarantee privacy, particularly for users from marginalized communities who might still perceive risks due to context collapse \citep{triggs-etal-2021-context}, and rely on ``throwaway accounts'' and other practices to negotiate identity boundaries \citep{leavitt-2015-this}. Furthermore, even with anonymous accounts, text posts encode stylistic markers that can reveal the identity of the author. Stylometry studies \cite{Holmes1998TheEO} suggest that such clues can help identify authors across multiple genres, domains, and discourse types \cite{Goswami_Sarkar_Rustagi_2009, Litvinova2020, markov-etal-2021-exploring}. %\mc{Do these references really support that point?}\cb{removed languages from the list, this was only included from the multilingual-based tasks but that is stretching it}

\begin{figure}
\centering
    \includegraphics[width=7.5cm]{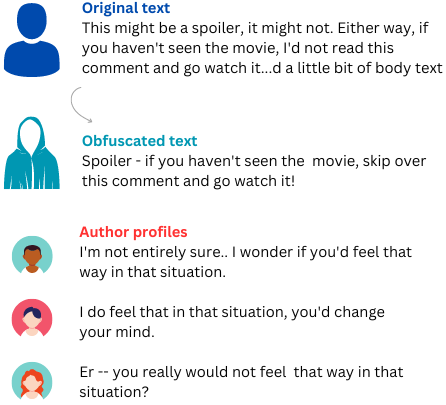}
    \caption{Authorship obfuscation as tested by attribution and verification attacks. A verification attack asks: Are the \textcolor{blue}{Original} and \textcolor{teal}{Obfuscated} texts written by the same author? An attribution attack asks: which author is the \textcolor{teal}{Obfuscated} text written by among a set of candidate authors, represented by their \textcolor{red}{author profiles}? }
    \label{fig:attributor_compared_to_verifier}
\end{figure}

Automatic authorship obfuscation holds the promise of helping people protect their privacy in online communication by automatically rewriting text to convey the original content while hiding the identity of the original author. Since there is little, if any, supervised training data for this task, existing approaches primarily rely on rule-based systems inspired by stylometry insights \cite{karadjov2017case}, repurposing machine translation systems for paraphrasing \cite{Keswani2016AuthorMT, shetty2018a4nt, altakrori-etal-2022-multifaceted}, or unsupervised style transfer models trained with dedicated adversarial objectives \cite{shetty2018a4nt, bo-etal-2021-er}. Other methods search for privatization-relevant surface edits using genetic search algorithms \cite{Mahmood2019AGH} or heuristic search \cite{bevendorff-etal-2019-heuristic}.
% \mc{only use the cite or citep command for citations that can be removed from the text without impact to fluency. for citations that are part of the text, use citet. There are multiple such issues throughout the paper.}
% which uses a genetic algorithm to search for suitable word replacements until an internal authorship model misclassifies, or \cite{bevendorff-etal-2019-heuristic}, which continuously makes minimal-cost edits until a divergence threshold between the privatization and the input is met. \mc{Suggested rewrite to get the gist of the techniques more directly: Other methods search for privatization-relevant surface edits using genetic search algorithms \citep{Mahmood2019AGH} or heuristic search \cite{bevendorff-etal-2019-heuristic}.} %which pushes towards an acceptable distributional distance (e.g. a threshold JSD) by imitating an authorial style using some defined character-level edits. \mc{the last sentence is jarring because there is too much technical detail for the intro, and also it does not make sense because it does not explain what kinds of search questions we are takking about and how they relate to privatization: explain the gist in simple terms instead.}  

Inspired by the ``Keep it Simple'' approach to unsupervised text simplification \citep{laban-etal-2021-keep}, we introduce ``Keep it Private'', an unsupervised authorship obfuscation technique based on large language models, which uses reinforcement learning to guide them to generate text that hides author identity while producing sound outputs that preserve the meaning of the original.  The impressive text generation abilities of language models suggest that they might help rewrite text in a way that is more natural and contextualized than stylometry-based edits. %Stylometry-based obfuscation is also inherently limited in the scope of edits performed (e.g., character and word-level edits), which makes it easy to target with authorship attributors \citep{uchendu2023attribution}. 
Large language models also offer an attractive general-purpose alternative to dedicated sequence-to-sequence models that rely on custom architectures, adversarial training, or parallel data such as \citet{emmery-etal-2018-style}. %\mc{wrong citation style. I will stop flagging those from now on, but do check the entire paper.}

To achieve privacy, an obfuscation model should be robust to any method that attempts to identify the author. Yet, prior work test on proxy tasks or against a single approach for authorship detection \citep{shetty2018a4nt, uchendu2023attribution, Mahmood2019AGH} in a small-scale authorship setting, with many writing samples per author. 
%\mc{The previous sentence is vague (lots of hedging with ``often'', ``sometimes'', and also has a single citation which is not enough to support what is presented as a description of a common trend in prior work. So need to add more citations here, and sharpen the writing.} 
To simulate a setting for users participating in online forums, we focus our domain on a large \texttt{REDDIT} dataset of 68k authors with short-medium length texts, and check to what degree authors remain private under our evaluation framework. We introduce a new evaluation framework for text obfuscation, where we privatize against several authorship attacks: automatic authorship verification and attribution, as delineated in ~\autoref{fig:attributor_compared_to_verifier}.  We introduce a method to guide LLMs to rewrite text for privatization via reinforcement learning, and show that our approach fools attribution and verification models the most, while maintaining soundness of outputs. We make available our scripts on GitHub\footnote{https://github.com/csbao/kip-privatization}.

\section{Background}\label{sec:background}

Our authorship obfuscation and evaluation strategies are informed by prior work on authorship analysis, which has been driven by the PAN shared tasks\footnote{https://pan.webis.de/shared-tasks.html} spanning profiling, attribution, style change detection, diarization, and obfuscation. We first review different ways of framing the adversarial task of author identification, before reviewing obfuscation methods themselves.

\paragraph{Authorship Identification} Identification can be framed either as \textbf{verification} \---\ the task of determining whether two texts were written by the same author \---\ or as \textbf{attribution} \---\ the task of identifying the author of a text among a set of authors represented by their writing samples.  %This can be done in a number of ways as shown by the variety developed at the PAN verification shared task \cite{bevendorff2020}.
%\mc{This is a little weak: what can be said about the salient dimensions of variations across all those implementations?}\cb{removing it for now, this is a point that is made more so in the results}
For either task, one key dimension of variation lies in the nature of writing samples provided, ranging from a single \textbf{instance} to an entire author \textbf{profile} which groups many instances written by a single user \cite{stamatatos2009}. In this work, we will evaluate our obfuscation techniques against a range of these types of adversaries.
%\mc{Don't we also consider instance based adversaries in the setting when we vary the profile length down to 1 comment?}\cb{yes, technically!}

%\mc{draw from Jack's document to make the point that identification is hard for humans, and that automatic techniques are often more successful at identification.} Automatic identification approaches were initially based on stylometric features such as $n$-gram occurrence patterns \cite{} \mc{n-gram tracing and other methods}. 

The PAN 2022 shared task on Authorship Verification \cite{stamatatos:2022} demonstrated that verification remains difficult in settings with varied domains and text lengths: many submissions were outperformed by a na\"ive baseline using cosine similarity of character $n$-gram representations of document pairs. We use this baseline as a verification adversary \texttt{VERIF\_CNG}.

Learning neural embeddings that represent authorship has recently proven effective in large data identification scenarios: In Learning Universal Authorship Representations (LUAR),  \citet{rivera-soto-etal-2021-learning} introduce embeddings trained contrastively to assign higher similarity to pairs of profiles by the same author than to pairs by different authors. %\mc{reworded previous sentence for flow.} 
The similarity score can be used for verification or for attribution. In attribution settings, ranking a list of candidate authors by LUAR similarity between profiles outperform stylometry-inspired approaches. We will use both approaches as identification adversaries in our evaluation.%\mc{TODO: add LUAR distance under verification in the main result table}

\paragraph{Authorship Obfuscation} 
The goal of obfuscation is to modify a document such that the obfuscated document cannot be traced to its original author. Many obfuscation techniques in the Author Masking series at PAN\footnote{\url{https://pan.webis.de/shared-tasks.html\#author-masking}} show that models trained for proxy rewriting tasks, such as round-trip translation \cite{Keswani2016AuthorMT, altakrori-etal-2022-multifaceted} can work well for masking authorship style. Some obfuscation models \cite{shetty2018a4nt, Mahmood2019AGH, bo-etal-2021-er} work in tandem with an adversary. Models designed explicitly to obfuscate stylometric features \cite{karadjov2017case, kacmarcik-gamon-2006-obfuscating} have been shown to fool identification models reliant on those features.
% More specifically targeting common features that these adversaries often use, models that explicitly obfuscate stylometric features \cite{karadjov2017case, kacmarcik-gamon-2006-obfuscating} have also been shown to fool identification models that use those features.\mc{Previous sentence is convoluted, need to simplify.}\cb{rewrote} 

% talk about similar task, but different methods + different data constraints (requires parallel data)
% \cb{Talk about same task, but different data/different methods}
Recently, end-to-end neural approaches that view obfuscation as a style transfer task have been proposed. \citet{bo-etal-2021-er} train a sequence-to-sequence model to generate text by masking the style from the input, without sacrificing aspects of fluency through its combination of a reconstruction loss and embedding reward at training time. \citet{emmery-etal-2018-style} train a sequence-to-sequence model on parallel data and an autoencoder on non-parallel data consisting of different editions of the English Bible, presumably written by different authors. %They show that the autoencoder outperforms the seq2seq model in a human evaluation. in these specific evaluation settings.  %, presumably due to limitations of the training data (due to factors of data size, limited authorial signal, etc.). 
%These methods show the current limitations of supervised deep learning for the obfuscation task, and beget training for a more low-resource setting, which is arguably more realistic when analyzing online text. 
These models show promise, yet require training from scratch using complex custom procedures.

Our approach mainly targets the unsupervised authorship obfuscation task in an open-world setting with much larger number of authors with limited writing samples per author. We approach this by fine-tuning general-purpose language models to produce meaning-preserving rewrites with masked authorial style. This work crucially trains for authorship obfuscation guided by a neural-based adversarial authorship embeddings \cite{rivera-soto-etal-2021-learning}, tested in a realistic online scenario.
\paragraph{Related Tasks}
% Contemporaneously to our work,\mc{that's not true anymore!} 
We discuss several tasks related to general obfuscation. Style imitation focuses on mirroring a target author's linguistic style. With a similar evaluation setup, \citet{Patel2022LowResourceAS} proposes a method rewriting \texttt{REDDIT} posts by prompting large language models (GPT-3 and BLOOM) to imitate the style of a target author \cite{rivera-soto-etal-2021-learning}.  However, we address the more general task of authorship obfuscation as opposed to impersonating a specific author. %\mc{Could talk about ethical considerations here. but low priority.}

Attribute obfuscation often pertains to altering text that is identifiable as a given attribute of the author, including gender and age. \citet{xu-etal-2019-privacy} introduce an approach to text rewriting by using reinforcement learning on top of round-trip MT to encourage rewrites that hide demographic attributes of the author. Meanwhile \citet{mireshghallah-berg-kirkpatrick-2021-style} propose a variational autoencoder technique that pools distinct styles associated with sensitive attributes to automatically rewrite text. 
\citet{shetty2018a4nt} present an unsupervised approach that adversarially trains a neural network to transfer text to protect sensitive attributes. While effective, style rewrites guided by a small number of coarse attributes are not well-suited to obfuscating authorship in online communities, given the large number of users organized within communities that are likely to share many manipulated attributes.

% \mc{I think a paragraph titled Related Tasks is called for, where you can briefly mention the Patel paper and also the attribute preservation tasks that came up in the reviews}\cb{Check!}

\section{Approach: The ``Keep it Private'' Model for Authorship Obfuscation }\label{kip_section}
Our approach to text privatization relies on large language models to rewrite the input text, and uses reinforcement learning to directly optimize metrics that encourage obfuscating the identity of the original author, while preserving the meaning and acceptability of the original text.  As a result, training is unsupervised and only requires examples of texts written by different authors, rather than supervised examples of obfuscation which are expensive to obtain at scale.

% \mc{Not sure whether this is doable before the deadline but studying the impact of the nature/distribution of training data on performance would be interesting}.\cb{Agreed! This is probably very important, with a focus on different domains or genres, maybe}

Our ``Keep it Private'' model (KiP) privatizes an input segment $X = (x_0, \dots, x_M)$ into an output $Y = (y_0, \dots, y_N)$ using a language model $p(y|x;\theta)$. Inspired by \citet{laban-etal-2021-keep}'s approach to unsupervised text simplification, we adopt their variant of Self-Critical Sequence Training ($k$-SCST). Just like the popular REINFORCE algorithm \cite{williams1992}, Self-Critical Sequence Training lets us optimize the gradient of the expected reward by sampling from the model during training, and treating those samples
as ground-truth labels weighted by the reward. Unlike REINFORCE, $k$-SCST relies on its own inference outputs to normalize the rewards observed. We optimize the following loss $\mathcal{L}$:
\begin{equation}\label{eq:kscsteq}
% \mathcal{R^S} = 
\mathcal{L} =  \sum_{j=1}^{k} (\overline{R^S} - R^{S_j}) \sum_{i=0}^{N}\log p(y_i^{S_j}|y_1^{S_j}...y_{i-1}^{S_j}, X)
\end{equation}

For each input $X$, we generate a set $S$ of $k$ output samples $Y^{S_j} = (y_0^{S_j}, \dots, y_N^{S_j})$. The loss  $\mathcal{L}$ weighs the log-likelihood of each sample $Y^{S_j}$ by the difference between $\overline{R^S}$, the mean reward over the $k$ samples and $R^{S_j}$, the reward of the current sample. The mean reward thus serves as a baseline to compare the individual sample reward, and yields a better estimate of the reward distribution. Minimizing the loss thus increases the likelihood of sample $S_j$ if it scores higher than the baseline mean reward.

Self-Critical Sequence Training was initially proposed for caption generation tasks \cite{rennie2017selfcritical}, and has also been used for other text generation tasks, including question generation \cite{zhang-bansal-2019-addressing} and summarization \cite{wang_etal_2018-abstract-summarization, celikyilmaz-etal-2018-deep}. \citet{laban-etal-2021-keep} showed the benefits of sampling $k$ candidate rewrites instead of one for unsupervised text simplification, a rewriting task that shares with obfuscation the need to rewrite stylistic attributes of the input text while preserving its meaning.

Within this framework, we consider a range of language models $P(Y|X;\theta)$ as base generators (\autoref{sec:base}), and design a set of rewards for authorship obfuscation (\autoref{sec:rewards}).

\subsection{Base Language Models}
\label{sec:base}

The loss (\autoref{eq:kscsteq}) can be used to optimize any language model $P(Y|X)$ that generates the output sequence $Y$ autoregressively from left to right. We consider two different language models:
\begin{enumerate*}
\item  \textbf{\texttt{GPT2}-medium} (345M parameters) \cite{Radford2019LanguageMA}, a decoder-only model,
% \item \textbf{\texttt{T5}} \cite{} an encoder-decoder model,
\item \textbf{\texttt{BART}-large} (406M parameters) \cite{lewis2019bart}, an encoder-decoder model.

\end{enumerate*}  We selected \texttt{GPT2} as our decoder-only model to align with the \cite{laban-etal-2021-keep} setting, and we select the \texttt{BART} models to investigate the impact of encoder-decoder models.
% \mc{KIP-T5 is also as a natural comparison point for KIP-DIPPER. We can probably get away with not including it if that's overaly complicated to include, since we have a lot of baselines, but we should at least know how to answer the question if asked when presenting the work.}\cb{The answer is that it didnt work as well as either GPT or BART when we were trying to explore other architectures. We could show that empirically but would need to do these experiments again.}
%\mc{there are many variants of these models, where applicable specify which one is used. Also we need to think about the order in which to present these models. What is the rationale for the current order?}\cb{It was initially ordered according to the order I tried them in :) But now I think it's okay to leave it as GPT2 , BART, and BART-para. The KiS implementation used GPT2.}

We consider two variants that encourage meaning preservation by fine-tuning them for paraphrasing tasks, namely: \begin{enumerate*}[resume]
% \item \textbf{\texttt{T5-para}} and
\item \textbf{\texttt{BART-para}} (406M parameters)  and
\item \textbf{\texttt{DIPPER-large}} (770M parameters)
\end{enumerate*}. \textbf{\texttt{BART-para}} is obtained by fine-tuning \textbf{\texttt{BART}} on meaning-preserving examples from 173.5k paraphrases coming from three datasets: QQP \cite{iyer-etal-2017-qqp} (149k pairs), PAWS \cite{zhang-etal-2019-paws} (21.8k pairs), and MSR \cite{dolan-brockett-2005-automatically} (2.7k pairs). \textbf{\texttt{BART-para}} was fine-tuned in a supervised, sequence-to-sequence fashion by generating one side of the pairs conditioned on the other over 4 training epochs. \textbf{\texttt{DIPPER}} is obtained by fine-tuning the \textbf{\texttt{T5-large}} \cite{raffel2023exploring} model on 152k pairs of synthetically perturbed aligned translations of non-English novels from the PAR3 dataset \cite{thai-etal-2022-exploring}. The perturbations allow the introduction of control codes at inference time to control the edit type and the intensity of edits made. %\mc{The previous sentence belongs earlier when discussing these 2 models}
%\mc{Need to discuss DIPPER!}
% on \mc{how many} supervised examples of meaning-preserving rewrites drawn from three paraphrase datasets \mc{name datasets, cite and explain what they contain.}. \mc{Also briefly explain fine-tuning procedure in one sentence at most.} 

%We vary the base generators to assess the impact of various architectures on the overall performance; namely, we consider a decoder-only model \texttt{GPT2} as well as an encoder-decoder models: \texttt{BART}. To improve the ability to sample more diverse, meaning-preserving outputs for our reinforcement learning, we experiment with a paraphrasing model fine-tuned on a corpus comprising MSR, QQP, and PAWS: \texttt{BART-Paraphrase}, and measure how effective this is at biasing the paraphrased samples away from the authorial style of the input. 

\subsection{Rewards}
\label{sec:rewards}

We design rewards that encode the three main desiderata of the text obfuscation task. A good rewrite should \textit{privatize} the text so that the identity of the author cannot be correctly detected, while being \textit{sound} (i.e., well-formed) and preserving the \textit{meaning} of the input. We describe how these are encoded in the reward below.% guardrails borrowed from \citet{laban-etal-2021-keep} to discourage pathological language model outputs. All rewards are computed using both the input and the generation.

\paragraph{Privacy} 
Among the many ways to quantify the privacy of a given text with attribution and verification tasks, we prioritize signals that are relatively cheap to compute as training rewards. We rely on the LUAR embeddings \cite{rivera-soto-etal-2021-learning} to compute cosine similarity $S_C$ of output $Y$ and input $X$ in the authorship embedding space, and subtract it from 1 to get the distance metric:
\begin{equation}\label{dist_privacy}
LUAR_{self} = 1 - C_S(L_X, L_Y).
\end{equation}
% \mc{What are $\hat{X}, \hat{Y}$? they have not been defined. is it just $X$ and $Y$? why a new notation?}

%\mc{what do the indices capture? are they needed here?}
% REMOVED.
% To disentangle content and style signals, we further reward outputs $Y$ that are distant from other texts written by the same author in authorship space, while normalizing for the distance between the two inputs considered as follows:  
% % \mc{how are these other texts selected at training time? how many are considered?}\cb{addressed}
% \begin{multline}\label{dist_privacy2}
% \begin{aligned}
% LUAR_{other_i} = dist(\hat{X_j}, \hat{Y_i}) - dist(\hat{X_i}, \hat{X_j})
% \end{aligned}
% \end{multline}
% Each example $X_i$ will only be paired with one other text $X_j$ selected randomly from the author's set of texts at training time.

%The first privacy reward, $LUAR_{self}$ (\autoref{dist_privacy}), simply uses the cosine distance between LUAR embeddings of the inputs and \texttt{KiP} generations. Intuitively, a higher cosine distance entails that a generation has better privatization. The second privacy reward, $LUAR_{other}$ (\autoref{dist_privacy2}), 
%scores with a positive author-contrastive setup in that the cosine distance is computed between the LUAR embeddings of a different text sample $X_j$ written by the same author as the current input $X_i$, as well as between the generation $Y_i$ of the current input and the other text sample $X_j$. We take the difference between the latter and the former to further disentangle signals of content and style within our privacy rewards. 

\paragraph{Meaning Preservation}\label{subsec:meaning_preservation} We compute the cosine similarity of \sbert embeddings \cite{reimers2019sentencebert} between the inputs and generations, as seen in \autoref{sim_sensibleness}.
\begin{equation}\label{sim_sensibleness}
SBERT_{self} = C_S(S_X, S_Y)
\end{equation}%\mc{same question: why the hats?}
To further improve saliency of the generations, we also retain the coverage model used in \citet{laban-etal-2020-summary, laban-etal-2021-keep} which is an informed gap-filling task on the original text, conditioned on the generation.

\paragraph{Soundness}\label{subsec:soundness} We use a grammatical acceptability model to assess whether an output is as sound as the input. Specifically, we use RoBeRTA-large fine-tuned on the CoLA dataset annotated with boolean grammatical acceptability labels \cite{warstadt-etal-2019-neural} to score both the input and output. The reward captures the agreement between these two soundness judgments:
\begin{equation}\label{cola_diff}
CoLa_{self} = CoLa(X) == CoLa(Y)
\end{equation}
% \mc{Is this really a strict equality test without any rounding or bucketing? that seems really hard to achieve in practice!} \cb{the output of CoLA here is boolean, not a scalar score. We convert to scalar score by averaging across snippets of X and Y }
To further improve soundness of the generations, we retain the fluency reward as described in \citet{laban-etal-2021-keep}. This reward works to maintain soundness in the base generator by using the model's likelihood score and a discriminator model trained adversarially.

\paragraph{Guardrails} Guardrails ensure that we keep generations aligned with basic rules, such as brevity and repetition. These are binary 0-1 values -- if triggered, they will effectively zero out the reward score, ensuring that the model does not learn from ``de-generations''.
We use the brevity guardrail from \citet{laban-etal-2021-keep} to penalize generations that fall outside of the 0.8--1.4 input-output length ratio range.%\mc{reworded the previous sentence, check that it is correct.}
The repetition guardrail discourages repetitions by penalizing outputs that contain any repeated 3-grams.%\mc{simplify:  The repetition guardrail discourage repetitions by penalizing outputs that contain repeated 3-grams.} 

%\mc{Need to briefly explain how a guardrail is different from a regular reward. that is not a standard ML concept.} 

\paragraph{Overall Reward Function} The final reward $R_S$ is the weighted logarithmic sum of the scalar components above, including the penalty guardrails:

% \begin{multline}
%   R_S = \log ({\gamma_1 \cdot LUAR_S + \gamma_2 \cdot LUAR_{S,P} + \\ \gamma_3 \cdot SBERT_{S} + \gamma_4 \cdot CoLa_{S} + \gamma_5 \cdot ChrF_{S} + \\ \gamma_6 \cdot Fluency_S})
% \end{multline}\label{combined_reward}

\begin{multline}
  R_S = \gamma_1 \cdot \log(LUAR_S) + \gamma_2 \cdot \log(SBERT_{S}) + \\ \gamma_3 \cdot \log(Fluency_S) + \gamma_4 \cdot \log(CoLA) \\ + \sum_{i=1}^{g} log(1-G_{i_S}) 
\end{multline}\label{combined_reward}

We use $\gamma_1 = 3, \gamma_2 = 2, \gamma_3 = 1, \gamma_4 = 1$, based on a small grid search in early experiments. %encoding our intuition that privacy should be weighted slightly higher than meaning preservation and soundness, and validated with a small grid search in initial experiments.

% \mc{What about the guardrails? no need to lsit them all, but they need to be accounted for somehow.}

% \begin{dmath}
%   R_S = \log ({\gamma_1 \cdot LUAR_S + \gamma_2 \cdot LUAR_{S,P} + \gamma_3 \cdot SBERT_{S} + \gamma_4 \cdot CoLa_{S} + \gamma_5 \cdot ChrF_{S} + \gamma_6 \cdot Fluency_S})

% \end{dmath}

% \begin{equation}\label{combined_reward}
% R_S = \gamma_1 \cdot LUAR_S + \gamma_2 \cdot LUAR_{S,P} + \gamma_3 \cdot SBERT_{S} + \gamma_4 \cdot CoLa_{S} + \gamma_5 \cdot ChrF_{S} + \gamma_6 \cdot Fluency(S)
% \end{equation}
% \mc{Missing the equation that defines the reward function on the basis of all these scores and also introduces the weights (since we will need to refer to those in the analysis, they need to be precisely defined)}

% \begin{enumerate}
% \item 
% \item 
% \end{enumerate}
% Put figure of the reward setup here
% Give outline of Section 3 KiS components here
\section{Experimental Design}
% \mc{If there is space, it is better form to have some text between the section and subsection.}
We describe our experimental design, including datasets, metrics, and models.
\subsection{Data}
\paragraph{Training} We base our training data on a corpus of English Reddit comments \cite{Baumgartner_Zannettou_Keegan_Squire_Blackburn_2020}. Our training split (\texttt{REDDIT}) is constrained to comments written by 30k authors, for a total of 7.12 million comments. Our data pipeline concatenates comments from one subreddit, written by the same author, into a pseudo-document until the pseudo-document reaches at least 250 words. We treat this as the author profile. For \texttt{KiP} models, we normalize by lowercasing, removing newline characters, duplicate spaces, and duplicate punctuation in order to encourage learning more substantial edits during the \texttt{KiP} process. 

\paragraph{Evaluation} 
We conduct our primary evaluation over the \texttt{REDDIT} dataset. Our data pipeline extracts 1600 ``needle'' comments from 100 authors (16 comments per author). In the \textbf{attribution} setting, these comments are grouped and concatenated by author to create an author profile, and then our adversarial authorship attribution model \texttt{LUAR} uses that needle profile to query for the most similar candidate profiles in our ``haystack'' candidate set of more than 1 million comments written from a superset of 68k authors (16 comments per author, wherein overlapping authors have disjoint comments from the ``needles''). In the \textbf{verification} setting, these comments are similarly grouped into the author profile, and then each profile is paired with a same-author profile constructed from a random sample of candidate texts, taken from the same author set. These pairs are provided to \texttt{VERIF\_CNG}, the strong character-based verification baseline from the PAN 2022 verification task \cite{stamatatos:2022}, in order to discriminate if they were written by the same author. \texttt{VERIF\_CNG} was trained on an even mixture of profile lengths (comments={1,2,4,8,16}) to be more robust to length size. Similar to the setup in \citet{rivera-soto-etal-2021-learning} and \citet{andrews-bishop-2019-learning}, our evaluation dataset is author-disjoint from our training data -- it contains comments by authors not present in the training data. %\mc{is that true only for verification or also for obfuscation? is the test set (authors) the same for both or not? this section is unclear. ALso need more statistics for the dev and test settings.} \cb{yes, the test set is the same in both verification and attribution and they use the same haystack to construct a profile, but by nature they are different b/c the attribution system uses a haystack, whereas the verification system uses a same-author profile constructed from the haystack. i wrote something up, hope that clears this up}

% \subsubsection{Fanfiction}
% Our corpus of Fanfiction \texttt{FANFIC} short stories comes from the PAN20 Verification Shared Task \cite{bevendorff2020}, crawled from fanfiction.net. This data was prepared by the organizers directly for authorship verification, so it consists of story pairs, with approximately half same-author and half different-author pairs. We use the \texttt{small} dataset here, amounting to almost 53k story pairs in total. For the obfuscation setting, our data pipeline chunks each story into 8-sentence paragraphs. If the chunk is less than 250 words, our pipeline concatenates it with the next chunk until it is. In the end, we have pseudo-documents consisting of at least 250 words, to be used as training input.
% This chunking is also done for our evaluation set up as well, but we do not hold a minimum requirement of 250 words there, to align with our setup for the Reddit data.

\subsection{Evaluation Metrics} \label{secautomatic_metrics}

We evaluate all outputs on the previously discussed aspects of privacy, meaning preservation, and soundness. 

\paragraph{Privacy} We evaluate privacy against a range of adversaries. We first report a LUAR distance, similar to the first privacy reward for our KiP model (Section~\ref{sec:rewards}), measuring distance between the LUAR embeddings of original inputs and privatized outputs. Second, we report the same retrieval metrics as in \citet{rivera-soto-etal-2021-learning} based on the LUAR authorship attribution model: (1) recall-at-8 (\texttt{R@8}) which checks whether the correct author appears amongst the top 8 authors predicted by the model, and (2) mean reciprocal rank (\texttt{MRR}), the average inverse rank of the ordered retrievals. We compute MRR as:
$ 1\over{n}$$\sum_{i}^{n}{1 \over RANK_{i}} $ over $n$ query author profiles. An MRR value of 1 indicates perfect retrieval, with the correct author always ranked first. An MRR of 0 indicates complete failure, with the correct author never retrieved.

% For these retrieval metrics, we introduce ``needles'', a collection of texts to be used as queries into a ``haystack'', a collection of texts written by a superset of the ``needle'' authors.\mc{Previous sentence is somewhat redundant with data description.}
For our \texttt{REDDIT} sample, the haystack is large, consisting of comments from 68k authors. The LUAR attributor performs well at this scale (93\% R@8, 83\% MRR) for the unmodified ``needles'' collection, which leaves room for a range of impact that various obfuscation models can have on the retrieval metrics after privatization.

Along with the retrieval metrics, we include the adversarial verification metric: c@1 score \cite{penas-rodrigo-2011-simple} of \texttt{VERIF\_CNG} to evaluate how often the obfuscation flips a testbed of entirely same-author profile pairs to the different-author label, while calibrating for model uncertainty.

\paragraph{Meaning Preservation} We compute self-\sbert, which is the cosine similarity of \sbert embeddings \cite{reimers2019sentencebert} from the input and the output in the test set to assess whether the privatized text preserves the meaning of the original. This is similar to how the respective reward is computed in \autoref{subsec:meaning_preservation}.

\paragraph{Soundness} 
For soundness, we compute CoLA-out -- the grammatical acceptability on the outputs. We also include \texttt{LenRatio} which computes the ratio of output to input character length as a soundness metric to spot-check that models are not producing unexpectedly short or long rewrites. %\mc{What is LenRatio exactly? Ratio of output to input length? based on which tokenization?}
% \mc{Missing explanation of soundness metrics. Perhaps simply reportint COLA-out is sufficient as it can be compared to the No-Op COLA score for context; Also I would add Length Ratio as a soundness metric: we are checking that the model does not go on extended ``rants'' in the outputs by comparing the length of the output to that of the input.}

\subsection{Conditions}

\paragraph{Baselines} 
We consider a diverse set of baselines:
\begin{itemize}
    \item the \textbf{Copy} baseline, which makes an exact copy of the input.
    \item a na\"ive \textbf{normalizer}, which replaces newlines with spaces, removes duplicate spaces, and removes duplicate punctuation. %\mc{any citation that supports this as a privatization model?}
    % \item \mc{The heuristic search model which ins in the table needs to be described}
    \item \textbf{round-trip MT (RTMT)}, an approach that paraphrases the input by repurposing an off-the-shelf m2m50 multilingual machine translation tool \cite{liu2020multilingual} to translate from English into German and back into English. 
    \item \textbf{LUAR-rescored RTMT}, which samples 4 generations per input and selects the best generation according to LUAR.
    %\mc{I would call this ``LUAR-rescored round-trip MT'' as ``extension'' doesn't carry much meaning}
    \item the \textbf{stylo} model, an obfuscation model that rewrites text to match pre-defined stylometric properties \cite{karadjov2017case}.  %\mc{"for some stylometric characteristics via some defined rules" is too vague to be useful. I would either be more specific or just say ``an obfuscation model that rewrites text to match pre-defined stylometric properties"} %\mc{it is not clear at all what it means for a model to be based on global averages. What is being edited? what is being averaged?} 
    \item prompting a \textbf{BLOOM-7b} \cite{workshop2023bloom} model to rewrite text into a neutral style. We use the prompts in step 1 of the target author imitation recipe from \citet{Patel2022LowResourceAS}.
\end{itemize}

We include other baselines to understand the impact of KiP: \textbf{\texttt{BART-Para}}, as described in \autoref{sec:base}, and \textbf{\texttt{DIPPER-large}} \cite{krishna2023paraphrasing}, which fine-tunes \texttt{T5-large} on the PAR3 dataset \cite{thai-etal-2022-exploring}, a collection of aligned literary translations. %\mc{replace or add DIPPER-L, also BART-para is not mentioned. Since the list of baselines is quite long, you can separate the baselines that represent prior work (ie everything up to BLOOM) from those that are meant to understand the impact of KIP (ie BART-Para and DIPPER-L)}
% \end{itemize}

\paragraph{Keep It Private Models} Our models are built as described in \autoref{kip_section}, resulting in four variants depending on the underlying base generator used: \texttt{KiP-GPT2, KiP-BART, KiP-BART-Para, KiP-DIPPER}. We used the same hyperparameters across base generators. We used Lamb optimizer \cite{you2020large} to optimize the model with learning rate 0.0001 on the loss function defined in \autoref{eq:kscsteq}. We use a training batch size of 4 inputs, sampling \emph{k}=8 runs per input for 8-SCST. Every training run is done on a single RTXA6000 GPU.

% \mc{describe trianing hyperparemeters, and any weights or settings that have not been described before}.

% \mc{Also I thought the result tables involves applies normalization to all outputs? Is that not the case? If so, it needs to be described.}\cb{described this in Section 4.1 (data section), let me know if this needs to be here as well.}\mc{It belongs in the data section only if it is applied to the data used by all models. If it is only used by KiP models that it is a pipeline approach that should be desribed here. However, this raises an issue about fairness of comparison: why isn't normalization used also for the stylo model, round-trup MT model, etc? is their performance artificially lowered by not using normalization?} \cb{Acknowledged. I'll try to add it to the other models tomorrow and re-eval.}

\section{Results}

\begin{table*}[htbp]
\small
    \begin{tabular}{  l  p{12cm}  }
        \toprule
\textbf{Model}      & \textbf{Output}   \\
\midrule 
\raisebox{-3ex}{Copy}
& And technically, looks like 50\% of <PERSON>'s comments are true and 25\% are half true. You could just as easily argue that 50\% of her comments are false (25\% false, 25\% half true). That's a deceptive way of looking at it.\\\hline\hline
 \addlinespace[0.1cm]
\raisebox{-2ex}{\texttt{Normalizer}}    &  \textcolor{orange}{a}nd technically, 
 looks like 50\% of <PERSON>'s comments are true and 25\% are half true. \textcolor{orange}{y}ou could just as easily argue that 50\% of her comments are false (25\% false, 25\% half true). \textcolor{orange}{t}hat's a deceptive way of looking at it. \\\hline

\raisebox{-2.5ex}{\texttt{RT MT}}   &  And technically, \textcolor{orange}{50\% of the comments from <PERSON> seem to be} true and 25\% half true. \textcolor{orange}{They} could equally easily argue that 50\% of \textcolor{orange}{their} comments are wrong (25\% \textcolor{orange}{wrong}, 25\% half true). \textcolor{orange}{This is} a \textcolor{orange}{seductive} way of looking at it. \\\hline
\raisebox{-5ex}{\texttt{Stylo}}    &  And technically, looks like \textcolor{orange}{Fifty percent of < Person > }'s comments are true and \textcolor{orange}{Twenty Five percent} are half true\textcolor{orange}{, and you} could just as easily argue that \textcolor{orange}{Fifty percent} of her comments are false ( \textcolor{orange}{Twenty Five percent false, Twenty Five percent} half true)\textcolor{orange}{, that is} a deceptive way of looking at it.  \\\hline
 \addlinespace[0.1cm]
 \raisebox{-1ex}{\texttt{BLOOM-7B}}    &  50\% of her comments are true \textcolor{orange}{AND} 25\% are half true. \textcolor{green}{Rewrite: 50\% of her comments are true AND 25\% are half true... (8 more repetitions).} \\\hline
 \addlinespace[0.1cm]
  \raisebox{-3ex}{\texttt{DIPPER 60,60}}    & \textcolor{orange}{logically}, looks like \textcolor{orange}{half of the person's} comments are true and \textcolor{orange}{half false}. You could just as easily argue that \textcolor{orange}{half} of her comments are false (25 false, half true). That's a deceptive way of looking at it.\\\hline\hline
 \addlinespace[0.1cm]

 \raisebox{-3ex}{\texttt{KiP-GPT2}}    &  And technically, looks like 50\% of <PERSON>'s comments are true and 25\% are half true. You \textcolor{green}{can do better than that and justify your stand by saying that 50\% of her comments are quite true and you shouldn't bother with testing. The problem is that while the majority of people who buy into your argument favour the slightly lower range of 25\% they are not practising anything wider than that and will usually settle for half true.}\\\hline

\raisebox{-2ex}{\texttt{KiP-BART}}    & And technically, \textcolor{orange}{it} looks like 50\% of \textcolor{orange}{the comments of <PERSON>} are true and 25\% are half true. \textcolor{magenta}{that's a deceptive way of looking it. y}ou could just as easily argue that 50\% of her comments are false (25\% false, 25\% half true ).  \\\hline

 \raisebox{-3ex}{\texttt{KiP-BART-Para}}    & And \textcolor{orange}{theoretically}, \textcolor{green}{it} looks like 50\% of the \textcolor{orange}{anonymous comments of someone} are true and 25\% are \textcolor{orange}{false}. \textcolor{magenta}{That's a deceptive way to look it}. \textcolor{orange}{I} could just \textcolor{orange}{the} easily argue 50\% of her comments false ( \textcolor{orange}{25 false, 25 true} ).\\\hline
 \raisebox{-3ex}{\texttt{KiP-DIPPER}}    & \textcolor{green}{-} And \textcolor{orange}{theoretically} \textcolor{green}{-}, \textcolor{green}{it} looks like \textcolor{orange}{haLF} of the \textcolor{orange}{comments of someone} \textcolor{orange}{A}re true and 25\% are \textcolor{orange}{false}. \textcolor{magenta}{That's a deceptive way to look at it}. \textcolor{orange}{I} could just as easily argue \textcolor{orange}{that half} of her comments \textcolor{green}{were} false.\\\hline

        \bottomrule
    \end{tabular}
        % \caption{Sample outputs. Rephrasings are italicized, additions are bolded, and reorderings are underlined. More outputs for a subset of these models can be found in \autoref{tab:more_obfuscation_outputs} in \autoref{apx:sample_outputs}.}\cb{Replaced color coding with bolds, italics, underlines in the table, though I think this is worse...}\mc{Then go  back to original color coding.} 
        \caption{Sample outputs. Rephrasings are colored in orange, additions are colored in green, and reorderings are colored in pink. More outputs for a subset of these models can be found in \autoref{tab:more_obfuscation_outputs} in \autoref{apx:sample_outputs}.}
        \label{tab:obfuscation_outputs} %\mc{I usually don't like this but given that we're so short on time, the font in this table could be made smaller}}

\end{table*}

\begin{table*}[ht]
\centering
\begin{tabularx}{\textwidth}{lrrrrrrr}%[table-format=2.6]}
\toprule
  & \multicolumn{4}{c}{\textbf{Privacy}} & \textbf{Meaning} & \multicolumn{2}{c}{\textbf{Soundness}}   \\ 
\cmidrule(lr){2-5}\cmidrule(lr){6-6}\cmidrule(lr){7-8}
  & \multicolumn{2}{c}{ \hlc[pink]{Attribution}} & \multicolumn{2}{c}{\hlc[cyan]{Verification}} & & & \\
& R@8 ↓ & MRR ↓ & CNG c@1 ↓ & LUAR ↑ & SBERT ↑ & COLA ↑ & LenRatio   \\  
  \midrule
\multicolumn{8}{l}{\emph{Baselines}}\\
~\texttt{Copy}  & 93.0 & 83.0  & 71.0 & 0 & \textbf{100.0} & \textbf{100.0} & 100.0 \\  
~\texttt{Normalizer} & 30.0 & 21.0  & 71.5 & 14.5 & 99.0 & 90.1  & 99.7 \\  
%~A* & 0.64 & 0.53  & 0.71 & 0.090 & \textbf{0.99} & 0.98 & 0.998 \\  
~\texttt{Stylo}  & 8.0 & 6.0  & 67.3 & 35.6 & 90.0 & 49.0 & 113.3 \\ 
~\texttt{Round-trip MT} & 56.0 & 43.0 & 71.8 & 17.7 & 89.0 & 85.5 & 100.4 \\  
~\texttt{LUAR-scored RTMT} & 51.0 & 38.0 & 73.5 & 23.0 & 84.4 & 86.4 & 101.07 \\  

% ~Bloom-3b & \textbf{5.0} & \textbf{3.0} & \textbf{63.8} & \textbf{55.7} & 40.5 & 83.3 & 2372.8 \\  
~\texttt{BLOOM-7b} & \textbf{3.0} & \textbf{2.0} & \textbf{56.8} & \textbf{46.9} & 44.3 & 94.2 & 2613.8 \\  
~\texttt{BART-Para} & 18.0 & 15.0 & 62.4 & 33.8 & 82.3  & 73.4 & 118.3 \\  
% ~\texttt{DIPPER-XXL 20L, 20O} & 14.0 & 9.0 & 71.8 & 36.0 & 82.9 & 78.0 & 102.5 \\  
% ~\texttt{DIPPER-XXL 60L, 60O} & 6.0 & 3.0 & 67.7 & 46.0 & 67.8 & 74.8 & 114.2 \\  
~\texttt{DIPPER 20L, 20O} & 16.0 & 12.0 & 73.2 & 38.6 & 84.5 & 78.3 & 112.3 \\  
~\texttt{DIPPER 60L, 60O} & 10.0 & 6.0 & 62.6 & 38.8 & 83.4 & 77.1 & 128.1 \\  

\midrule
\multicolumn{8}{l}{\emph{Keep It Private Models}}\\
~\texttt{KiP-GPT2} & 9.0 & 6.0 & 72.8 & 39.7 & \textbf{78.6} & \textbf{76.7} & 478.2 \\ 
~\texttt{KiP-BART} & 11.0 & 9.0 & 55.9 & 35.4 & 73.5 & 66.7 & 146.3 \\  
~\texttt{KiP-BART-Para}  & 4.0 & 4.0 & 46.6 & 42.6 & 77.1 & 61.5 & 89.7 \\
~\texttt{KiP-DIPPER}  & \textbf{2.0} & \textbf{2.0} & \textbf{42.3} & \textbf{45.2} & 70.1 & 63.3 & 93.4 \\

\bottomrule
\end{tabularx}
\caption{Obfuscation performance over the \texttt{REDDIT} evaluation dataset. Keep It Private Models generally improve the privacy of generated text compared to baselines, but the improvement is more consistent against an attribution than a verification adversary. Privacy improvements come at the cost of a small degradation in meaning preservation and soundness.}  %\mc{TODO: fix scores to have a constant number of decimals and correct alignment within a column; pick a better name for Heuristic Search; decide whether we need both MRR and R@8; bold top scores in each column based on statistical significance tests, or possibly top score within each column and each half of the table (baseline vs. KiP); also what is the COLA score? It looks like it's the cola-diff score for some rows but the cola out score for others??? need to pick whichever makes KiP-BART-para look good :); finally make sure that revise caption once table is finalized to make sure that the take-aways are correct.}}
    \label{tab:reddit_results}
\end{table*}

% \autoref{tab:panResults} shows our evaluation scenario of our models benchmarked against the described metrics on the \texttt{PAN} dataset. The \texttt{RED} evaluation will be shown as well. The normalizer retains the most sense while achieving limited privacy in both the attribution and verification metrics. 
\paragraph{Overview} We present the main evaluation results on the \texttt{REDDIT} evaluation dataset in \autoref{tab:reddit_results}. We set the upper bound for our obfuscation methods by highlighting the performance of the \textbf{Copy} baseline, which represents the performance of adversarial authorship tools on the text as-is.
% \mc{Need to explain what is different between the 2 Tables!} \cb{We set the upper bound for our obfuscation methods by highlighting the performance of the \textbf{Copy} baseline, which represents the performance of adversarial authorship tools on the text as-is.}
%privacy evaluation thus indicates the performance of our authorship attribution and verification adversaries in standard settings, which sets an upper bound for our obfuscation methods. 
The high attribution scores suggest that LUAR embeddings provide sound adversaries in these settings, and we compare that with our baselines and proposed KiP models. Finally, we discuss our human evaluation to further validate paraphrasing adherence of some of our obfuscation methods, and investigate how author profile length is a factor in the privacy evaluation.
%\mc{Why introduce the copy baseline here? I would introduce it together with the other baselines, and briefly mention the implications here. Although it is not clear why some baselines are discussed in this paragraph and why there is a separate paragraph called baselines next. Need to clarify the structure.}\cb{COMMENT - agreed, i'll move the description of the Copy baseline to the Baselines in 4.3, but still want to highlight the implications of the baseline performance (basically, the reader should see a higher delta performance against the Copy baseline as a positive indication of obfuscation)}
% \mc{An overview paragraph placed here should be an overview of the entire section, not just the first 2 subsections. SO either split the current section into 2 (main results and analysis), or give a more global overview.}

\subsection{Baselines} 
Overall, our set of \textbf{baselines} show that the effectiveness of existing obfuscation methods varies widely. Many of the baselines preserve meaning well based on the high SBERT scores, however this is simply a consequence of limited or formulaic edits. The trivial edits of the Normalizer surprisingly degrade authorship attribution more than round-trip MT, with only a slight improvement when selecting the best-performing sample according to \texttt{LUAR}. The baseline that privatizes best is the \texttt{Bloom-7b} model, however, this comes at a heavy meaning cost. Qualitatively, this method also frequently produced repetitions, leading to outputs that are on average 20 times longer than inputs. On the other hand, the \texttt{Stylo} model also privatizes well, but it creates unsound outputs that appear unnatural to the human eye\footnote{Sample model outputs can be found in \autoref{tab:obfuscation_outputs}}, as we will see in the human evaluation (\autoref{subsec:analysis}). We also see strong performance from the paraphrasing baselines -- \texttt{DIPPER} specifically shows strong privacy performance with higher meaning preservation depending on edit control code. The verification performance of all baselines except \texttt{Bloom-7b, DIPPER 60L, 60O}, and \texttt{BART-Para} remain close to that of the Copy control.

\subsection{KiP models} 
The \texttt{KiP} models generally improve the privacy metrics over the baselines by performing edits that are more effective at fooling attribution and verification adversaries. As can be expected, this makes it harder to preserve meaning or soundness, although the \sbert scores remain high (70 or above). \texttt{KiP-GPT2} ranks well on privacy of attribution and meaning preservation metrics, but fails on soundness with outputs that are on average up to 4.78 times longer than inputs. %\mc{Where does the 6 number come from? I only see the 478 length ratio and don't understand the connection to 6}\cb{thanks for catching!}
Manual inspection shows that, despite the guardrails, \texttt{KiP-GPT2} is prone to hallucinating long tangents that are fluent (as observed by high \texttt{CoLA} scores) but only topically connected to the original text, as illustrated in \autoref{tab:obfuscation_outputs}. This fools LUAR-based adversaries, but not \texttt{VERIF\_{CNG}}. Using the \texttt{BART} language model as an underlying generator curbs this pathological behavior. The \texttt{KiP-BART model} improves privatization against both attribution and verification adversaries compared to \texttt{KiP-GPT2}, with more sound outputs.

Using a base generator fine-tuned for paraphrasing (\texttt{KiP-BART-Para, KiP-DIPPER}) results in more extensive edits that achieve improved privatization performance without hurting meaning preservation compared to the more conservative \texttt{KiP-BART} and even the aggressive baseline \texttt{DIPPER} model, however at a cost in acceptability.  Overall, this shows that the \texttt{KiP} paraphrasing models address the actual task of text obfuscation compared to baselines and improves on privatization, including against adversaries that are independent from the privatization rewards it was trained on. 

% \mc{missing a discussion of Kip-DIPPER!}

% \mc{missing a discussion of BLOG results and how they relate to REDDIT results.}

%Though privacy and meaning metrics are more stable across the KiP models compared to the baselines, we observe a wide range in soundness metrics. This is especially salient in KiP-GPT2, where outputs are on average 6 times longer than inputs. 

%We see KiP-BART-para improve in all privacy metrics over the other KiP-generator variants, while keeping meaning relatively stable (especially when comparing KiP-BART-para with its KiP-BART counterpart). Although the KiP-BART* models fail to retain as much sense as the GPT2 model, they seem more tethered to the input, as shown by the LenRatio. Because obfuscation is a length-invariant task, the presence of much longer or shorter outputs is undesirable. Though we see rather low COLA metrics in the \texttt{KiP-BART*} models on \autoref{tab:reddit_results}, we see that it is possible to significantly improve them with weight selection.

Finally, privatization models as a whole tend to be significantly more successful at fooling attribution than verification models, confirming the need to measure progress against a diverse range of adversaries. Though target use cases might differ, we argue that an effective privatization method should fool all or many detectors.

\begin{figure}[h!]
\centering
    \includegraphics[width=7.5cm, height=5.5cm]{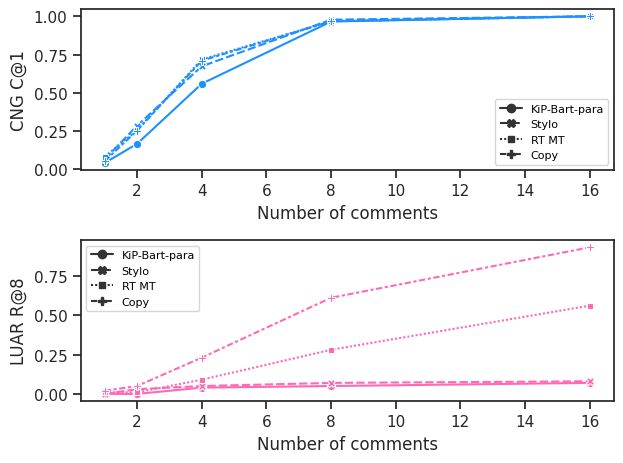}
    \caption{Higher values on the Y-axis indicate better performance of the adversarial model, and thus, worse performance in the obfuscation. A subset of baselines (\texttt{Stylo, RT MT, Copy}) is compared against the \texttt{KiP-Bart-Para} model over the first five powers-of-2 progressions for author profile size: 1 $\rightarrow$ 16 comments.}
    \label{fig:attributor_x_verifier}
\end{figure}

% \mc{The structure of this section is confusing; I suggest you add another subsection here called Analysis or something as it seems that you are transitioning from the discussion of the results in Table 2 (and 3?) to something else.}
\subsection{Analysis}\label{subsec:analysis}

We include additional evaluation assessing how author profile length impacts the privacy evaluation. We also report a human evaluation to further validate the meaning preserving qualities of a subset of our systems.

\begin{figure}[h!]
% \centering
    \includegraphics[scale=0.32]{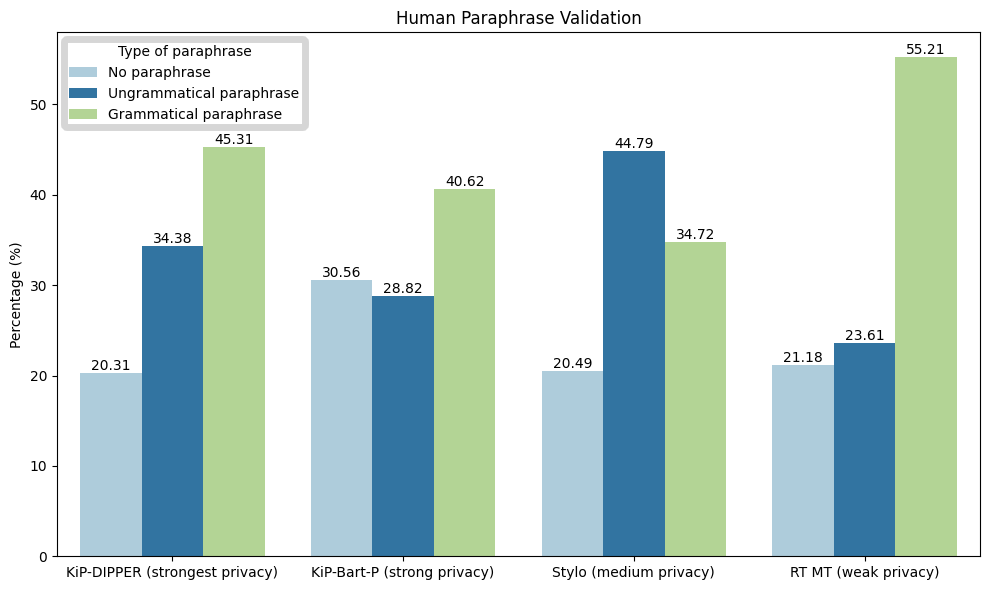}
    \caption{Results from a crowdsourced paraphrase pair evaluation. Systems are ordered from strongest (left) to weakest (right) in automatic privacy performance. Meanwhile, \texttt{KiP-DIPPER} produces more grammatical paraphrases than the other models, validating KiP-DIPPER's rewriting promise for achieving privacy and meaning preservation jointly.}

    \label{fig:ParaphraseValidation}
\end{figure}

\paragraph{Human Evaluation}
Because our goal is to prevent automatic authorship identification, we test the privacy-preserving aspects of system outputs according to an extensive automatic evaluation. To complement the automatic meaning preservation and soundness metrics described in \autoref{secautomatic_metrics}, we focus on validating generation quality of our proposed system in the human evaluation. We include 99 system outputs from the \texttt{REDDIT} evaluation each from \texttt{Stylo, RT MT, KiP-Bart-Para, KiP-DIPPER}, totalling 396 outputs. Instead of asking annotators to separately assess meaning preservation and fluency, we adopt the three-point Likert scale as done by \citet{hallinan2023steer, iyyer-etal-2018-adversarial} for \textbf{paraphrase validation}: (0 = no paraphrase, 1 = ungrammatical paraphrase, 2 = grammatical paraphrase). We recruited 36 English-fluent participants using Prolific, and compensated them at rates complying with local wage standards. Results suggest that \texttt{KiP-DIPPER} outputs are meaningful and well-formed paraphrases of the original texts according to humans (Figure~\ref{fig:ParaphraseValidation}), despite being obfuscated with strong privacy-preserving edits.

% \mc{This evaluation should include Kip-DIPPER (it does not matter what was submitted and what was added later, we need to prsent the whole thing together as a coherent whole).}

\paragraph{Author Profile Length} \autoref{fig:attributor_x_verifier} shows the performance of our obfuscation model in the \hlc[cyan]{verification} and \hlc[pink]{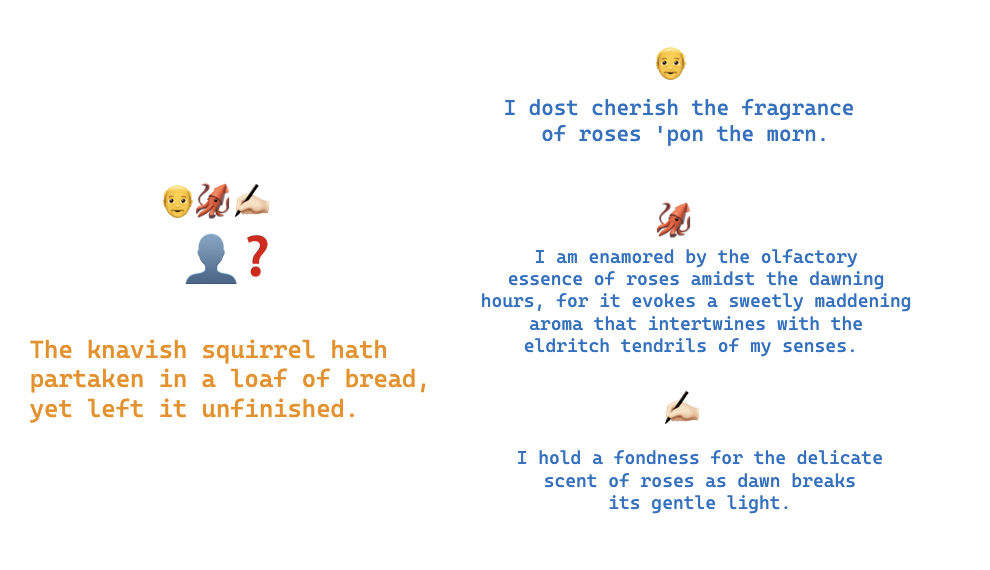} scenarios over several \# of \texttt{REDDIT} comments in author profiles. %As in \autoref{tab:reddit_results}, to create author profiles, we group comments according to author and then concatenate only the selected number. 
In the verification setting, we create a problem set containing on one side, the concatenations of the selected number of the obfuscations of the ``needles'' set, and on the other, unmodified versions of the ``candidates'' set written by the same authors as in the ``needles'' set. \texttt{VERIF\_CNG} was trained on an even mixture of lengths (comments=\{1,2,4,8,16\}) in author profiles to provide a lens into various author profile sizes in the test set. In the attribution setting, we apply the same control on length per author profile, using the entire pool of unmodified ``candidates'' written by 68k authors. We limit our analysis here to the \texttt{KiP-Bart-para} model and three baselines.

Performance of the \texttt{KiP-Bart-Para} model worsens slowly in the \hlc[pink]{attribution setting} (0\% R@8 at 1 comment $\rightarrow$ 4\% R@8 by 16 modified comments versus 2\% R@8 at 1 comment $\rightarrow$ 93\% R@8 by 16 unmodified comments). We observe a similar trend for Copy in the \hlc[cyan]{verification setting}: \texttt{VERIF\_CNG} is not able to discriminate obfuscations from same-author pairs until a certain profile length, at which point the \texttt{VERIF\_CNG} can correctly discern the pair as same-author. Though there is a gap in favor of the obfuscation at 2 and 4 comments, this gap is closed by the time the author profile grows to 8 comments. \texttt{KiP-Bart-para} is unable to fool \texttt{VERIF\_CNG} at longer author profiles.

%We see closer gaps between the no-op baseline and \texttt{KiP-Bart-Paraphraser} as author profiles grow. %This ability decreases slightly at longer text lengths \cb{i think due to topic or content drift which LUAR can handle well, but our verifier is not LUAR-based}, but we still see closer gaps between the no-op baseline and the model as author profiles grow. 

\paragraph{Reward Ablations}
Ablation experiments confirm that the meaning preservation, soundness and privacy components of the reward are all necessary. We provide the detailed results in ~Appendix \ref{appendix:ablation_table}.

\paragraph{Out of Domain Evaluation} We complement the REDDIT evaluation results by reporting additional results on the \texttt{BLOG} evaluation dataset in \autoref{appendix:blog_evaluation} \autoref{tab:blog_results}. Here, we confirm that the KiP models improve performance against both privacy adversaries. However, we see a larger reduction in meaning preservation with the \texttt{DIPPER} model, compared to \texttt{BART-Para}.   %\mc{Explain takeaway from the BLOG results.}

\section{Conclusion}

We introduced a method for training obfuscation models that use reinforcement learning on top of pre-trained language models to obfuscate the authorship of the original text. This method relies on a diverse range of rewards, crucially including neural authorship representations to judge authorship signals. The resulting outputs are edited more substantially than with existing obfuscation baselines, thereby improving privacy, while preserving meaning and soundness better than other successful obfuscation strategies. We find that using paraphraser base models lead to better balancing of both privacy and meaning preservation in the resulting KiP models. Additionally, conducting an extensive evaluation with diverse adversaries and input lengths highlights some important performance differences --- namely, that it is difficult to fool verification systems with longer obfuscated author profiles, even if they fool attribution systems. This calls for more research into designing robust evaluation benchmarks for obfuscation systems, to assess and catch failure cases that can map to different real-world scenarios. %We note this need also across domains, discourse types and languages.

% \mc{an important take-away that is not represented in the conclusion is that using paraphrasers as a base model for KiP achieves the best trade-offs of privatization and meaning preservation.}

\section{Acknowledgments}
This research is supported in part by the Office of the Director of National Intelligence (ODNI), Intelligence Advanced Research Projects Activity (IARPA), via the HIATUS Program contract \#2022-22072200006. The views and conclusions contained herein are those of the authors and should not be interpreted as necessarily representing the official policies, either expressed or implied, of ODNI, IARPA, or the U.S. Government. The U.S. Government is authorized to reproduce and distribute reprints for governmental purposes notwithstanding any copyright annotation therein.

% \mc{Reiterate key findings for the paper clinic}

% Here, point to real-life implementation differences in verification vs attribution. Fooling an attributor model can easily be done in cases where we e.g. break soundness, but a verification model can easily pick up on these things (see figure 2). However, verification does not attribute texts (obviously), just spot-checks when a text has been tampered with given appropriate authorship profile. a motivating scenario might be in sensitive scenarios when you have been in communication with author A (e.g. a spy) and need to verify if correspondence was actually written by author A, and need to check if text has been tampered with. An attribution setup may also be too expensive to configure, and attribution depends a lot on coverage)
% Future work -- on the modelling side, we should investigate the impact of training data (different cuts: subreddits, temporals, maybe?) on the RL process in KiP to seee how robust it is. On the evaluation side, we should expand to varying haystacks in an open-world setting to see how this changes metrics. We could also consider different privacy rewards.

\section*{Limitations}
Our primary evaluation is limited to two English datasets on short to medium-length texts. Because we only require data annotated by author ID, this method should be able to easily port to new datasets or new domains (for example, written fanfiction) in principle. However, this needs to be validated in a broader range of settings, especially because many of the reward components use English models. Furthermore, we focused exclusively on \textbf{\textit{automatic}} authorship attribution and verification, and did not explore how people with varying expertise in authorship analysis might manually assess the resulting texts.

%scaling to new languages would require nontrivial renovation of the existing reward set.
% \mc{Experiments are limited to one dataset, one language.}

\section*{Ethics Statement}

Our work illustrates an improvement on automated obfuscation software, novelly applying a fine-tuning strategy for the task of general authorship obfuscation. The overarching goal for technologies that enable obfuscation on text data is to protect attribution of individuals or groups in cases where authorship metadata is scrubbed (e.g. using a pseudonym) \cite{afsaneh-2021-why}. At the same time, powerful obfuscation tools could be used to threaten, cyberbully, or otherwise endanger other individuals without accountability or fear of retribution.

Additionally, though this work does not explicitly target author style imitation, we acknowledge that obfuscation can cause existing identification tools to mis-attribute authorship to unsuspecting people. Because these identification tools have shown to be incredibly accurate for authorship retrieval, in scenarios where these users are unaware that texts have been modified, mis-attribution is a serious concern \citep{altakrori-etal-2022-multifaceted}. 

The Reddit dataset used as training data in our work is licensed as CC-BY-4.0 protocol, which stipulates that publicly released data will be used exclusively for research purposes. All pre-trained models used in this work are publicly available on HuggingFace\footnote{https://huggingface.co/}, and we ensured that the research methodologies described in this work are aligned with licensing permissions. All methods and models described in this work are for research purposes, and are not intended for commercial use.

We manually reviewed our evaluation data to ensure that PII or personal entity identifiers were masked. We did not scrub swear words. For the human evaluation, we manually reviewed outputs and removed any candidates with hate speech.
% \mc{Overall, we stress that this work serves as a proof of concept of the appropriateness of the KiP framework for authorship obfuscation, and do not recommend its use in sensitive settings. (SHould we way something like that or not? I could go either way)}\cb{I like it, but not sure if it contradicts the point about journalists posting without fear of retribution from governments and such.. this is mostly a "don't let people apply it misanthropically" concern i guess, which is probably true for a lot of modern-day NLP}

 % Scientific work published at EMNLP 2023 must comply with the \href{https://www.aclweb.org/portal/content/acl-code-ethics}{ACL Ethics Policy}. We encourage all authors to include an explicit ethics statement on the broader impact of the work, or other ethical considerations after the conclusion but before the references. The ethics statement will not count toward the page limit (8 pages for long, 4 pages for short papers).

% \section*{Acknowledgements}


% \mc{There are 15 papers in the references that are cited without any venue. Many of those have been published, need to update the reference accordingly.}
\bibliography{anthology,custom,hiatus}
\bibliographystyle{acl_natbib}

\section{Appendix}\label{sec:appendix}
\begin{appendices}
% \section{Reward Weight Ablation}
% \paragraph{Reward Design} To better understand trade-offs across evaluation dimensions, we measure the impact of reward weights on the results (\autoref{tab:reward_ablation}).  As expected, we see that upweighting LUAR-based rewards compared to SBERT (e.g. rows 1 and 5, compared to rows 3 and 7) leads to better LUAR-based evaluation at the expense of SBERT and for \texttt{BART-para}, CoLA, and vice versa (rows 3 and 7, compared to rows 1 and 5), which leads to better SBERT scores at the expense of LUAR. This suggests that finding the right balance of these rewards is an important direction for future work, having established that KiP provides a good framework to address authorship obfuscation.
\section{BLOG Evaluation Results}
We start with 400 ``needle'' \texttt{BLOG} \cite{schler2006effects} snippets from 200 authors (2 comments per author). In our evaluation set, there are 211 words per \texttt{BLOG} snippet with 2 snippets per author, versus ~37 words per \texttt{REDDIT} comment with 16 comments per author. The attribution evaluation for this data is as described in the \texttt{REDDIT} setting. We include the \texttt{BLOG} dataset in our evaluation to show that the proposed models operate well out-of-domain, and on slightly longer text lengths. Results are reported in \autoref{tab:blog_results}.
\label{appendix:blog_evaluation}

For our \texttt{BLOG} sample, the haystack is smaller than that of the \texttt{REDDIT} sample. We observe that the LUAR attributor performs well out-of-domain at 51\% R@8, 41\% MRR, also leaving room to assess impact of the included obfuscation methods.

\begin{table*}[ht]
\centering
\begin{tabularx}{\textwidth}{lrrrrrrr}%[table-format=2.6]}
\toprule
  & \multicolumn{4}{c}{\textbf{Privacy}} & \textbf{Meaning} & \multicolumn{2}{c}{\textbf{Soundness}}   \\ 
\cmidrule(lr){2-5}\cmidrule(lr){6-6}\cmidrule(lr){7-8}
  & \multicolumn{2}{c}{ \hlc[pink]{Attribution}} & \multicolumn{2}{c}{\hlc[cyan]{Verification}} & & & \\
& R@8 ↓ & MRR ↓ & CNG c@1 ↓ & LUAR ↑ & SBERT ↑ & COLA ↑ & LenRatio   \\  
  \midrule
\multicolumn{8}{l}{\emph{Baselines}}\\
~ Copy  & 51.0 & 41.0  & 89.4 & 0.0 & 100.0 & 100.0 & 100.0 \\  
~\texttt{Normalizer} & 14.0 & 12.0  & 89.4 & 19.1 & 99.4 & 84.2 & 98.2  \\  
%~A* & 0.64 & 0.53  & 0.71 & 0.090 & \textbf{0.99} & 0.98 & 0.998 \\  
~\texttt{Stylo}  & 10.0 & 6.0  & \textbf{90.6} & 25.6 & 91.8 & 34.0 & 109.4 \\  
~\texttt{Round-trip MT} & 34.0 & 23.0  & 89.3 & 13.4 & \textbf{92.9} & 84.0 &  98.8\\   
~\texttt{LUAR-scored RTMT} & 32.0 & 23.0 & 88.9 & 15.5 & 90.7 & \textbf{85.3} & 98.9 \\  

% ~Bloom-3b & \textbf{5.0} & \textbf{3.0} & \textbf{63.8} & \textbf{55.7} & 40.5 & 83.3 & 2372.8 \\  
~\texttt{BART-Para} & \textbf{8.0 }& 6.0  & 85.3 & \textbf{35.0} & 76.5 & 62.0 & 71.4 \\   
% ~\texttt{DIPPER-XXL 20L, 20O} & 10.0 & 7.0  & 83.1 & 30.7 & 84.7 & 77.7 & 88.4 \\   
% ~\texttt{DIPPER-XXL 60L, 60O} & \textbf{6.0} & \textbf{4.0}  & 82.9 & \textbf{38.9} & 72.7 & 78.3 & 89.1 \\  
~\texttt{DIPPER 20L, 20O} & 12.0 & 10.0 & 89.5 & 31.3 & 92.3 & 76.4 & 108.5 \\  

~\texttt{DIPPER 60L, 60O} & 10.0 & \textbf{5.0} & 89.7 & 33.9 & 90.6 & 74.5 & 110.2 \\  
\midrule
\multicolumn{8}{l}{\emph{Keep It Private Models}}\\
% ~\texttt{KiP-GPT2} & - & -  & - & - & - & - & - \\  
% ~\texttt{KiP-BART} & - & -  & - & - & - & - & - \\  
~\texttt{KiP-BART-Para} & 4.0 & 5.0 & 87.6 & 38.9 & \textbf{69.9} & \textbf{66.0} & 126.3 \\  
~\texttt{KiP-DIPPER} & \textbf{1.0} & \textbf{0.0} & \textbf{86.1} & \textbf{44.7} & 64.9 & 63.5 & 163.9 \\  

\bottomrule
\end{tabularx}
\caption{Confirming the same findings from \autoref{tab:reddit_results} on the out-of-domain \texttt{BLOG} dataset, we observe strong attribution performance from the baseline paraphrasers \texttt{BART-Para} and \texttt{DIPPER}, and stronger attribution performance from the KiP models. However, we see limited defense against the \texttt{VERIF\_CNG} attack in this domain.}\label{tab:blog_results}
\end{table*}

\section{Human Evaluation Setup}
As it is difficult for humans to identify authorship signals in natural language texts, we focused our human evaluation on assessing \textbf{paraphrase validity} to ensure quality of system generations.

\begin{figure*}[ht!]
\centering
    \includegraphics[scale=0.4]{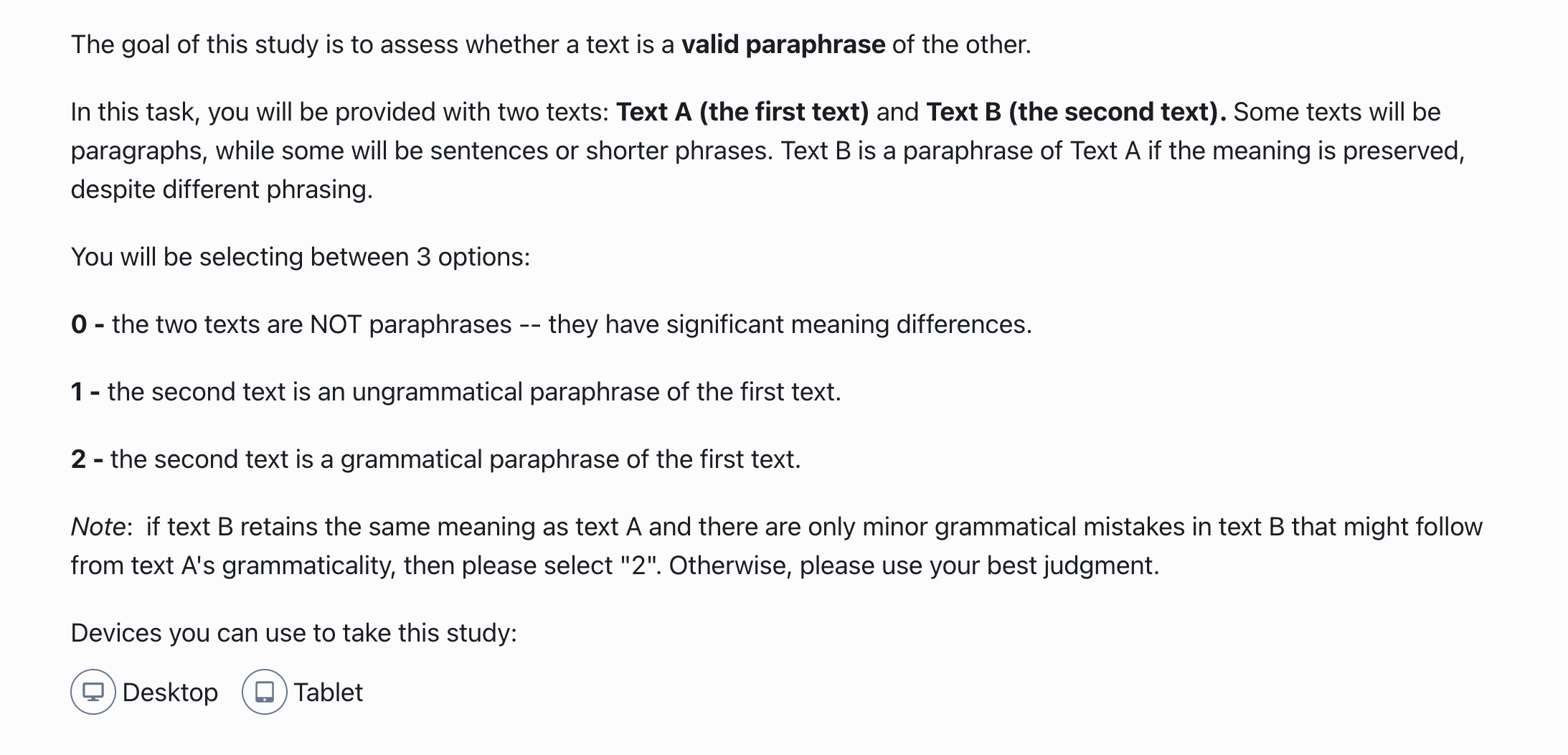}
    \caption{Instructions given to study participants. 27 English-fluent participants were recruited via Prolific.}
    \label{fig:instructions_human_eval}
\end{figure*}

\begin{figure}[h!]
\centering
    \includegraphics[scale=0.35]{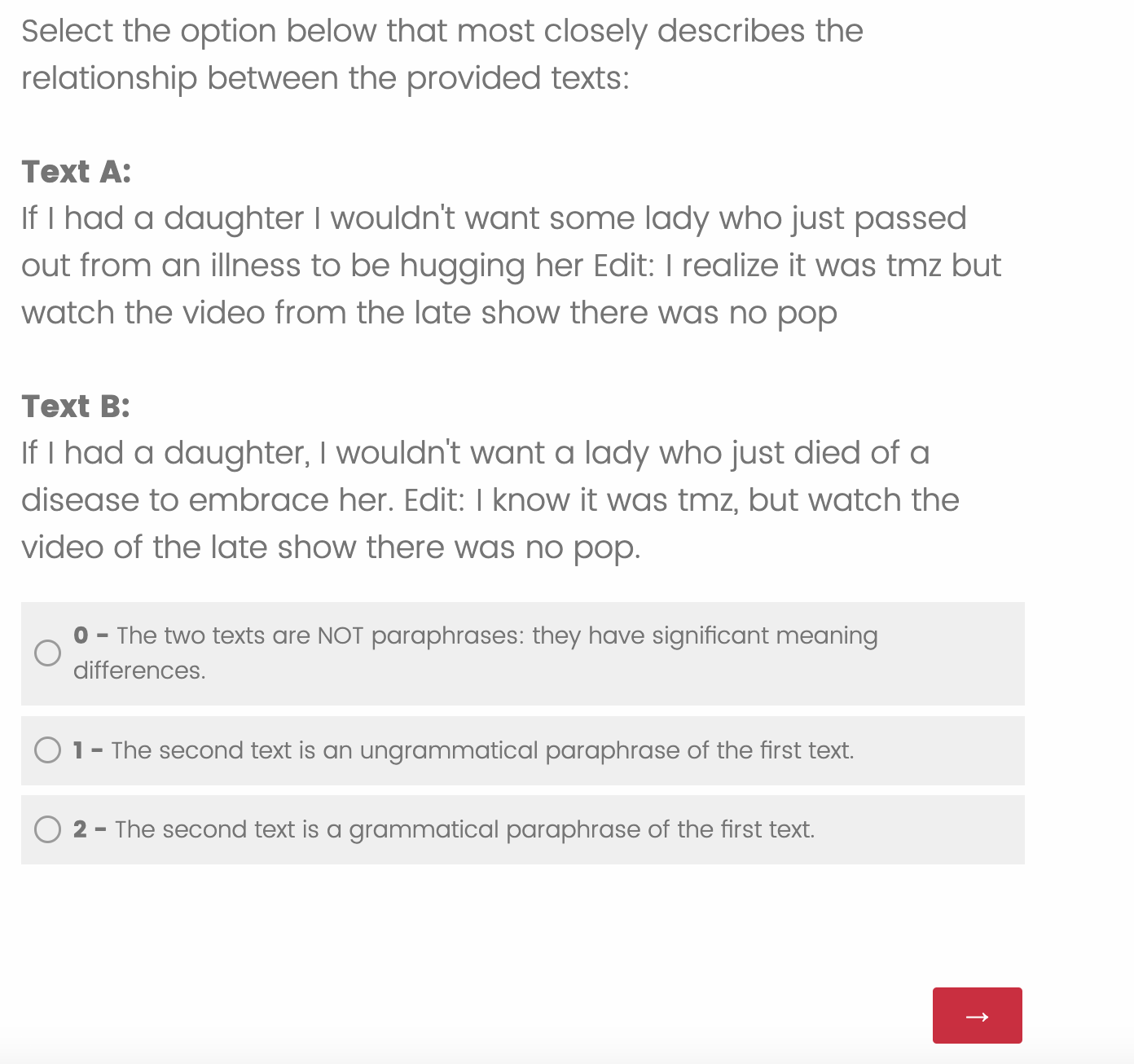}
    \caption{A sample multiple-choice question given to annotators.}
    \label{fig:q1}
\end{figure}

\begin{figure}[h!]
\centering
    \includegraphics[scale=0.35]{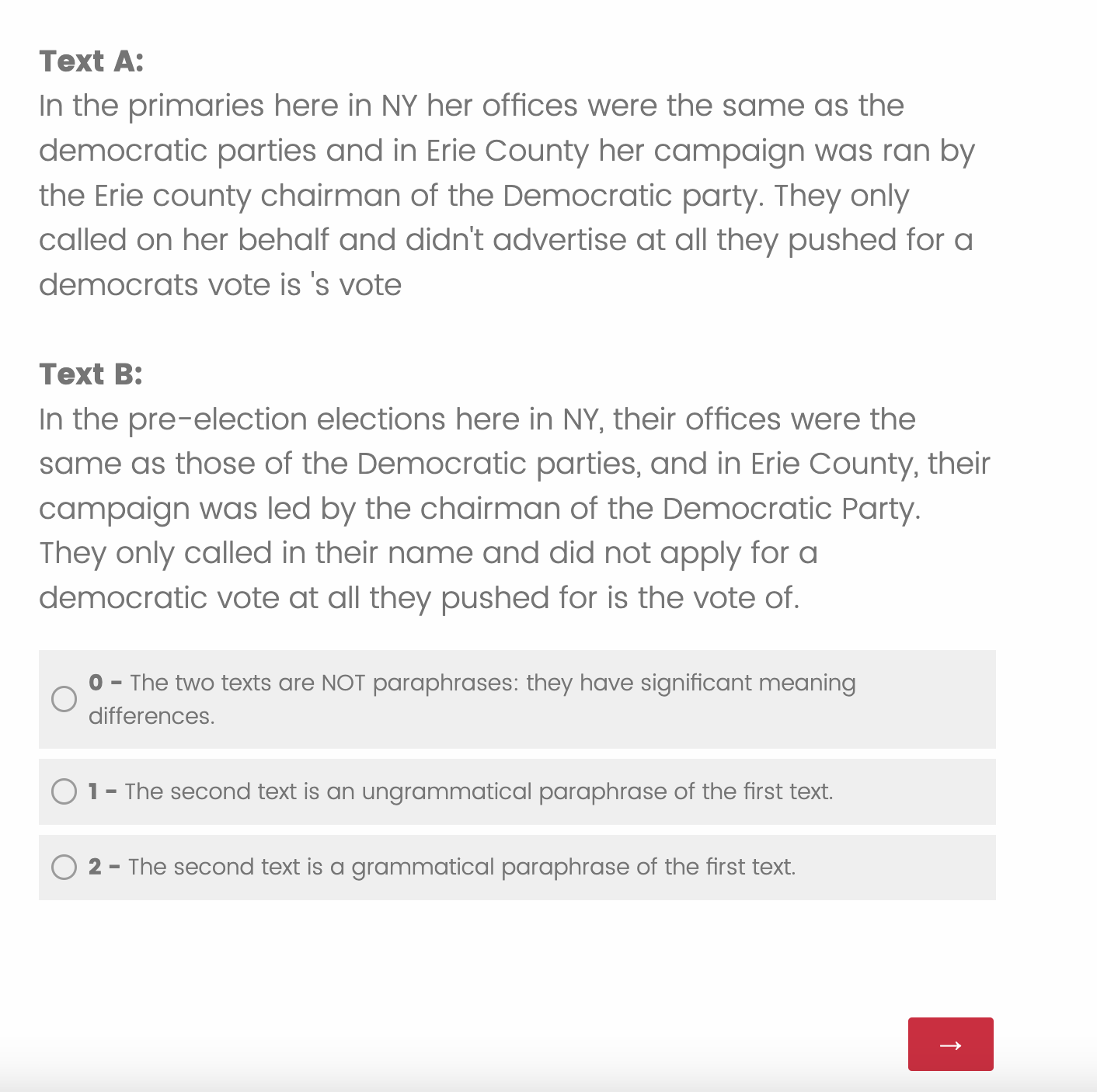}
    \caption{Another sample multiple-choice question given to annotators.}
    \label{fig:q2}
\end{figure}

\autoref{fig:instructions_human_eval} shows the annotation instructions, and \autoref{fig:q1} and \autoref{fig:q2} show sample annotation screens. We recruited in total 27 participants. Each participant was paid complying with local wage standards. Each input-output pairing was annotated 3 times, and we took the majority vote as the score for the pairing. 

\section{Reward Component Ablation}
\paragraph{Reward Components} To better understand trade-offs across evaluation dimensions, we measure the impact of reward components on the results (\autoref{tab:component_ablation}).  We confirm that the meaning preservation, soundness and privacy components of the reward are all necessary. As expected, when we remove the LUAR-based reward leads to better privacy evaluation at the expense of SBERT and for \texttt{BART-para}, CoLA, and vice versa (rows 3 and 7 compared to rows 1 and 5), which leads to better SBERT scores at the expense of LUAR. 
\label{appendix:ablation_table}
\captionsetup{width=.75\textwidth}
\begin{table*}[p]
\centering
\begin{tabularx}{0.5\textwidth}{l @{\makebox[1.5em][r]} | llllrrr}
    
% {@{}llrrrrrrrrr@{}} % repeats {c|} 18 times
\toprule

  \multicolumn{1}{l}{\textit{Model}} & \multicolumn{1}{l}{LUAR} & \multicolumn{1}{l}{SBERT} & \multicolumn{1}{r}{CoLa}  \\  
  \hline\hline
% \multirow{5}{*}{\texttt{GPT2}} & $\gamma_1=\textbf{4},\gamma_2=3,\gamma_3=2$ & \textbf{0.6187} & 0.518 & 0.817\\
% & $\gamma_1=\textbf{4},\gamma_2=3,\gamma_3=\textbf{4}$ & 0.417 & 0.8 & 0.66 \\
% & $\gamma_1=0,\gamma_2=2,\gamma_3=\textbf{5}$ & 0.527 & 0.621 & 0.881 \\
% & $\gamma_1=2,\gamma_2=2,\gamma_3=2$ & 0.562 & \textbf{0.624} & \textbf{0.895} \\
% \hline
 Remove guardrails & 0.632 & 0.378 & 0.383 \\
Remove fluency & 0.246 & 0.818 & 0.608 \\
Remove meaning & 0.265 & 0.677 & 0.756 \\
Remove privacy & 0.384 & 0.812 & 0.591 \\
~\texttt{KiP-BART-para} & 0.426 & 0.771 & 0.615 \\
\hline
\bottomrule
 \end{tabularx}\caption{Impact of removing various components on LUAR distance, SBERT similarity and CoLA, an acceptability judgment.}
  \label{tab:component_ablation}

\end{table*}

\section{BLOOM Prompts}
As with all language models used in this work, we pulled language models from model repositories on the Hugging Face\footnote{https://huggingface.co/} platform. For our prompting experiments, we used BLOOM-7b \footnote{https://huggingface.co/bigscience/bloom-7b1}. We followed the procedures listed out in \cite{Patel2022LowResourceAS} to prompt BLOOM into rewriting the given \texttt{REDDIT} comments in a neutral style. We set default decoding parameters $\texttt{temperature}=0.7, top_p=1.0$, and prompt with:

\begin{quote}
Passage: \textcolor{green}{\{Input Reddit comment\}} \\
Paraphrase the passage in a simple neutral style.\\
Rewrite: 
\end{quote}
and allow BLOOM to generate until the <eos> token. We treat the model generation as the obfuscated text. We did not do any decoding parameter experiments.

\section{Sample of obfuscation outputs}\label{apx:sample_outputs}
More sample outputs can be found in \autoref{tab:more_obfuscation_outputs}.

\begin{table*}[h]

    \begin{tabular}{  l  p{12cm}  }
        \toprule
\textbf{Model}      & \textbf{Output}   \\
\midrule 
\raisebox{-8ex}{Copy}
& \begin{enumerate}
    \item It's all that sustains her at this point.
    \item It's more of a reflection of the low enthusiasm of her supporters. People simply aren't excited about <PERSON>. Obviously it doesn't mean much in the real world you can't be so literal
    \item  He's the nimblest of all navigators
    % \item You sure? As as rep told me new notes are not going out till 19th anywhere in the world. And it's official shop.
    \item For ios 100\% no, the permission settings is still saying that it tracks your movement only when the app is open.
\end{enumerate}\\\hline\hline
\raisebox{-12ex}{\texttt{RT MT}}    &  \begin{enumerate}
\item It is all that it supports at this time
\item It is more a reflection of the low enthusiasm of their supporters. People are simply not enthusiastic about <PERSON>. Obviously it doesn't mean much in the real world you can't be so literal
\item He is the cleverest of all navigators
% \item Are you sure? As Rep told me, new scores will not go up to the 19th anywhere in the world. And it is an official shop.
\item For ios 100\% No, the permission settings still say that it only tracks your movements when the app is open.
\end{enumerate} \\\hline

\raisebox{-12ex}{\texttt{Stylo}}    &  \begin{enumerate}
\item It 's all that lengthen or extend in duration or space her at this point
\item It 's more in a reflection in the low enthusiasm in her supporters, people simply are n't excited about < Person >; and obviously it does n't mean much in the true international you ca not be so literal
\item "He 's the nimblest in all the member of an aircrew who is responsible for the aircraft 's course
% \item Good, You sure; yet as as rep told me new notes are not going out till Nineteenth any part of the globe, and and well, it 's official shop.
\item For ios One hundred percent no, the permission settings is still saying that it tracks your movement only when the app is open.
\end{enumerate} \\\hline\hline
 \addlinespace[0.1cm]
 \raisebox{-12ex}{\texttt{KiP-BART}}    &  \begin{enumerate}
 \item it's all that sustains her at this point
 \item it's more of a reflection of the low enthusiasm of her supporters. people simply aren't excited about <PERSON>. obviously it doesn't mean much in the real world you can't be so literal
 \item he's the nimblest of all navigators
  % \item you sure? as as rep told me new notes are not going out till 19th anywhere in the world. and it's official shop.
 \item for ios 100\% no, the permission settings is still saying that it tracks your movement only when the app is open. 
 \end{enumerate}\\\hline
\raisebox{-12ex}{\texttt{KiP-BART-para}}     &  \begin{enumerate}
\item At this point, it's all that sustains her.
\item It's more of a manifestation of low expectations from her supporters. people simply aren't excited about about about 'PERSON'. it doesn't mean much in the world, can be literal
\item He is the nimblest of all navigation leaders.
% \item replanned note are not going to the 19th is anywhere in world. You sure?
\item For ios 100\% no. the permission settings says it only tracks Movement when The app Open
\end{enumerate} \\\hline\hline

 \raisebox{-12ex}{\texttt{KiP-DIPPER}}    &  \begin{enumerate}
 \item All that she has to support her at this point is this
 \item This simply shows the low enthusiasm of her supporters. Obviously, you can't be so literal in the real world. They are simply not excited about her.
 \item He is the nimblest of all navigators
  % \item you sure? as as rep told me new notes are not going out till 19th anywhere in the world. and it's official shop.
 \item No, Ios is still saying it only tracks your movements when the app is open.
 \end{enumerate}\\\hline\hline

\\\bottomrule
    \end{tabular}
        \caption{More privatized Reddit comments from a subset of the explored obfuscation methods.}\label{tab:more_obfuscation_outputs}
\end{table*}

\end{appendices}

\end{document}